\documentclass{llncs}
\usepackage{graphicx}
\usepackage{epstopdf}
\usepackage{subfigure}

\usepackage{tikz}
\usetikzlibrary{arrows,positioning,shapes.geometric}
\begin{document}
\title{Domain knowledge assisted cyst segmentation in OCT retinal images}

\author{Karthik Gopinath \and Jayanthi Sivaswamy}
\institute{RIAG, India.}

\maketitle

\begin{abstract}
3D imaging modalities are becoming increasingly popular and relevant in retinal imaging owing to their effectiveness in highlighting structures in sub-retinal layers. OCT is one such modality which has great importance in the context of analysis of cystoid structures in subretinal layers. Signal to noise ratio(SNR) of the images obtained from OCT is less and hence automated and accurate determination of cystoid structures from OCT is a challenging task. We propose an automated method for detecting/segmenting cysts in 3D OCT volumes. The proposed method is biologically inspired and fast aided by the domain-knowledge about the cystoid structures. An ensemble learning method- Random forests is learnt for classification of detected region into cyst region. The method achieves detection and segmentation in a unified setting. We believe the proposed approach with further improvements can be a promising starting point for more robust approach.  This method is validated against the training set achieves a mean dice coefficient of 0.3893 with a standard deviation of 0.2987 

\end{abstract}

\section{Introduction}
Optical coherence tomography (OCT) is a non-invasive imaging modality that captures 3D projections of the retinal layers using  low-coherence light waves \cite{costa2006retinal}. In assessing retinal diseases, OCT is a handy tool. It has a potential to provide 3D information and hence analyze subretinal layers which are not captured by more conventional techniques such as color fundus imaging. Early versions of OCT were of limited use owing to low image quality and acquisition time. Fourier transformation of the optical spectrum of the low-coherence interferometry refers to Spectral-domain OCT (SD-OCT) or Fourier-domain OCT. SD-OCT is less noisy and significantly faster than the previous technology. The increased speed and number of scans renders greater image detail and clarity \cite{schuman2008spectral}. The higher SNR of SD-OCT allows assessment of smaller pathological changes. SD-OCT is becoming one of the most important ancillary test for the diagnosis of sight degrading diseases today. 3D cross sectional volumetric images of the retina and the sub-retinal layers are imaged from SD-OCT to detect pathologies such as cysts in addition to retinal pathologies. 
 
Cystoid macular edema (CME) \cite{rotsos2008cystoid} \cite{scholl2010pathophysiology} represents a common pathological appearance in retina and co-occurs in a variety of pathological conditions such as intra ocular inflammation, central or branch retinal vein occlusion \cite{hayreh1994retinal}, diabetic retinopathy \cite{browning2008optical} and most commonly following cataract extraction \cite{gass1969follow} . CME appearing in one eye, increases the likelihood of similar appearance in the second eye. CME causes, multiple cyst-like (cystoid) regions filled with fluid appearing in macula. Cysts cause retinal swelling or edema. The areas of retina in which the cells are displaced are cysts. CME leads to many complications such as loss of vision. Persistent CME for more than 6 to 9 months leads to chronic macular changes, with permanent impairment of central vision. 

While automated detection of CME in SD-OCT images is relevant for early diagnosis and prevention of vision-loss, this task is nontrivial and challenging. SD-OCT images often suffer from noise, apart from the demarcation between subretinal layers not being very clear. While long and high power exposure to LASER can alleviate some of these difficulties, patient health and other medical considerations often impose some necessary bounds on such approaches. Thus automated CME detection from noisy SD-OCT data is a challenging and relevant problem in retinal image analysis

This work proposes a biologically inspired and simple methodology for an automatic segmentation of candidate cyst regions, in SD-OCT retinal images. This paper is organized as follows: in Section 2 the methodology for automatic cyst detection is exposed. Section 3 shows the results obtained with this technique and in Section 4 conclusions and future lines of work. 
\section{Methodology}
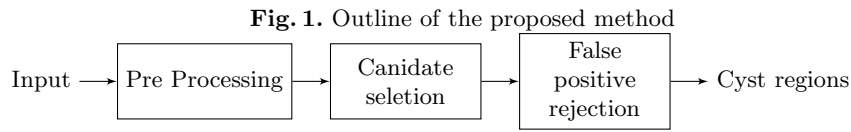
\begin{center}
\begin{figure}
\caption{Outline of the proposed method}
\label{Fig:Block diagram }
    \begin{tikzpicture}[>=latex']
    
        \tikzset{block/.style= {draw, rectangle, align=center,minimum width=2cm,minimum height=1cm}}
        \node (input_img) {Input};
        \node [block, right =5mm of input_img]  (start) {Pre Processing};
        \node [block, right =5mm of start] (acquire) {Canidate \\ seletion};
        \node [block, right =5mm of acquire] (rgb2gray) {False\\ positive\\ rejection};
        \node [right =5mm of rgb2gray](Segmented) {Cyst regions};

        \path[draw,->] (input_img) edge (start)
					(start) edge (acquire)
                    (acquire) edge (rgb2gray)
                     (rgb2gray) edge (Segmented)

                    ;
                    
    \end{tikzpicture}
    \end{figure}
\end{center}

The detection or segmentation of cyst candidate regions is a complex task. Our method to accomplish this task is composed of different stages (Fig. \ref{Fig:Block diagram }) a preprocessing stage: to perform size normalizations of individual slices, A total variance denoising, enhancement using center surround; a candidate selection phase: segmentation of ILM and RPE layers to reduce the search region, MSER features for region selection; finally, false positive rejection stage. 

\subsection{Preprocessing}
The training data has 15 SD-OCT volumes from 4 different scanners. The number of slices across the training set differ from 49 to 200. The size of each slice is neither constant for an individual scanner nor across scanners. For further processing, the size of the image is standardised for all volumes. All the scanned volumes are resized to a 512X256 pixels.
\\SD-OCT images are degraded by the speckle noise. Speckle noise is signal dependent and the pattern of the speckle depends on the structure of the imaging tissue.The speckle noise is a multiplicative noise and can be modelled using Rayleigh distribution. Speckle corrupted SD-OCT images has a have high total variation i.e. the integral of the absolute gradient of these images is high. We use total variational denoising \cite{boyd2004convex} technique which has advantages over the traditional denoising methods such as linear smoothing and median filtering. The Total Variational approach will reduce the texture content resulting in a smooth piecewise constant images preserving the edges \cite{yang1995optimal} \cite{hamza1999removing} \cite{buades2005review}. 
\\Center-surround difference is a biologically inspired technique of finding local extrema in images \cite{itti1998model}. As a saliency mechanism, its effectiveness has been demonstrated in a variety of applications. Since cyst region candidates are locally dark, we use center-surround difference in a multi-scale setting to detect the locally minimum regions as cyst candidates.
\subsection{Candidate selection}
Selection of region of interest is a vital task as it helps in reducing the search region for identifying cysts and computation time. Region of interest for cyst detection task are the retinal layers  between   Internal Limiting Membrane (ILM) and Retinal pigment epithelium (RPE). Areas like vitreous cavity and area below choroid layers are not of interests in detection of cyst regions.Various segmentation algorithms have been proposed for segmentaion of retinal layers. Methods like pixel intensity variation based ILM and RPE segmentation\cite{fabritius2009automated}, active contour with a two-step kernel-based optimization scheme \cite{garvin2008intraretinal}, complex diffusion filtering with combined structure tensor replacing thresholding \cite{cabrera2005automated}. In this work we use a conceptually simpler, yet accurate, graph theory based segmentation approach \cite{chiu2010automatic}.
\\Candidate regions are selected using Maximally stable extremal regions(MSER). This feature computation gives a set of  stable extremal region in an image.  This feature is used to detect the multi-scale objects without any smoothing involved. Both fine and large structure can be detected using this feature.

\subsection{False positive rejection}
The regions picked from the MSER features contain both cyst and non cyst regions. A bounding box around the candidate region is taken as the input for feature extraction. A local descriptor based on the texture of pattern \cite{medathati2010local} is calculated for the patch. This feature is extracted for all detected regions and a random forest with 50 trees is trained for the classification task. The detected cyst regions is considered as the segmentation result and the performance of the system is evaluated.

\section{Experimental Results and Discussion}
The performance of our method is evaluated using OPTIMA Cyst Segmentation Challenge MICCAI 2015 training dataset. The dataset includes 15 volumes of SD-OCT scans from  4 different scanners containing a wide variety of retinal cysts. The details of the dataset is shown in Table \ref{Table:dataset}
\begin{table}
\begin{center}
\caption{Dataset used}
\label{Table:dataset}
\begin{tabular}{|p{60 pt}|p{60 pt}|p{60 pt}|p{60 pt}|p{60 pt}|p{60 pt}|p{30 pt}|p{30 pt}|p{30 pt}|}
\hline
Scanner &\textbf{Spectralis}&\textbf{Cirrus}&\textbf{Topcon}&\textbf{Nidek}&\textbf{Total}\\ \hline 
Training &\textbf{4}&\textbf{4}&\textbf{4}&\textbf{3}&\textbf{15}\\ \hline 

\end{tabular}
\end{center}
\end{table}

Segmentation results are quantitatively measured using Dice coefficient (DC) defined below. 
\begin{equation}
Dice \  coefficient=2{\frac{|Detected \cap  GT| }{|Detected| + |GT|}}
\end{equation}

The maximum possible value for DC is 1(indicating a perfect match between the result and Ground Truth).

Validation on the training set is performed using a leave-one-out cross validation approach. The mean and standard deviation for all the volumes with ground truth taken as union of Graders 1 and 2 is mentioned in Table \ref{Table:Dice} below.
\begin{table}
\begin{center}
\caption{Dice coefficient with GT as union of two graders}
\label{Table:Dice}
\begin{tabular}{|l|l|l|l|l|l|l|l|l|}

\hline
 Dice coefficient 
 &\multicolumn{1}{c|}{\textbf{Spectralis}}
 &\multicolumn{1}{c|}{\textbf{Cirrus}}
 &\multicolumn{1}{c|}{\textbf{Topcon}}
 &\multicolumn{1}{c|}{\textbf{Nidek}}\\
\hline
Mean  &\textbf{0.3024}&\textbf{0.3728}&\textbf{0.3736}&\textbf{0.4613}\\ \hline
Maximum  &\textbf{0.8596}&\textbf{0.9246}&\textbf{0.8638}&\textbf{0.8367}\\ \hline
Standard deviation &\textbf{0.3436}&\textbf{0.2510}&\textbf{0.2849}&\textbf{0.2823}\\ \hline

\end{tabular}
\end{center}
\end{table}

The DC is zero for a slice when no cyst region is detected. The mean DC for a volume is lowered when there are few slices with zero cyst detection. The mean DC values for the given volumes across 3 scanners is tabulated in Table \ref{Table:Dice_nozero}.

\begin{table}
\begin{center}
\caption{Dice coefficient excluding zero DC slices}
\label{Table:Dice_nozero}
\begin{tabular}{|l|l|l|l|l|l|l|l|l|}

\hline
 Dice coefficient 
 &\multicolumn{1}{c|}{\textbf{Spectralis}}
 &\multicolumn{1}{c|}{\textbf{Cirrus}}
 &\multicolumn{1}{c|}{\textbf{Topcon}}
 &\multicolumn{1}{c|}{\textbf{Nidek}}\\
\hline
Mean  &\textbf{0.6652}&\textbf{0.5606}&\textbf{0.5357}&\textbf{0.5957}\\ \hline
Maximum  &\textbf{0.8596}&\textbf{0.9246}&\textbf{0.8638}&\textbf{0.8367}\\ \hline
Standard deviation &\textbf{0.1195}&\textbf{0.1851}&\textbf{0.1754}&\textbf{0.1496}\\ \hline

\end{tabular}
\end{center}
\end{table}

The analysis is discussed module-wise. Table \ref{Table:Small_cysts} indicates the percentage of small ($<$200 pixels), medium (200 to 2000 pixels) and large($>$2000 pixels) sized cysts detected. The SNR was observed to vary across the four scanners. Detection fails due to noise, largely for small sized-cysts and since they constitute a large proportion of the cysts the effect is an deterioration the net DC value. Specifically, the MSER feature is unable to detect many small cyst regions due to the presence of noise.  

\begin{table}
\begin{center}
\caption{Detection of cysts based on size from MSER features}
\label{Table:Small_cysts}
\begin{tabular}{|l|l|l|}

\hline
 Size of cysts &\multicolumn{1}{c|}{\textbf{No. Cyst Present}}&\multicolumn{1}{c|}{\textbf{Percent detected correctly}}\\ \hline
Small  &\textbf{2514}&\textbf{61.76}\\ \hline
Medium  &\textbf{2109}&\textbf{80.02}\\ \hline
Large &\textbf{243}&\textbf{88.04}\\ \hline
\end{tabular}
\end{center}
\end{table}

Along with the cyst regions, there are a lot of non-cyst regions picked up by the candidate selection stage. These constitute false positives. The supervised classifier will aid in reducing these false positives. The random forest ensemble classifier performance with respect to cysts of different sizes is listed below in Table \ref{Table:forest_cysts}. It is apparent that the detection is once again worst for small-sized cysts compared to medium and large ones due to the aggressive rejection by the random forest stage.

\begin{table}
\begin{center}
\caption{Detection of cysts after random forest based on size}
\label{Table:forest_cysts}
\begin{tabular}{|l|l|}

\hline
 Size of cysts &\multicolumn{1}{c|}{\textbf{Percent detected correctly}}\\ \hline
Small  &\textbf{27.83}\\ \hline
Medium  &\textbf{51.18}\\ \hline
Large &\textbf{76.10}\\ \hline
\end{tabular}
\end{center}
\end{table}

\section{Conclusion}
This paper presented a method to detect and segment the retinal cysts in a SD-OCT volume. 
The failure in detection of small cyst regions (indistinguishable from noise) is a major contributor to poor performance. Other contribution is not having an explicit segmentation stage. The MSER regions are often only partially overlapping with true cyst regions which leads to a lowering of DC value of even true detections. This was found to be the case in medium-sized cysts. 
Based on the above observations, future investigations can adopt different detection strategies depending on the size of the cysts in the SD-OCT volumes to improve the results. An advanced denoising algorithm might also help in segmenting small sized regions. Inclusion of a final segmentation stage (post rejection) will also aid in enhancing the performance.


\bibliographystyle{splncs}
\bibliography{ver_4}

\end{document}